\documentclass[11pt]{article}
\usepackage[T1]{fontenc}
\usepackage[margin=1in]{geometry}
\usepackage{amsmath,amssymb}
\usepackage{graphicx}
\usepackage{booktabs}
\usepackage[table]{xcolor}
\usepackage{pifont}
\usepackage{cite}
\usepackage[hidelinks]{hyperref}

\newcommand{\cmark}{\ding{51}}
\newcommand{\xmark}{\ding{55}}

\title{Localization Lens for Improving Medical Vision-Language Models}
\author{
    Hasan Farooq\textsuperscript{1} \quad
    Murtaza Taj\textsuperscript{1} \quad
    Mehwish Nasim\textsuperscript{2} \quad
    Arif Mahmood\textsuperscript{3}\\[0.8em]
    \small\textsuperscript{1}Lahore University of Management Sciences, Lahore, Pakistan\\
    \small\textsuperscript{2}The University of Western Australia, Perth, Australia\\
    \small\textsuperscript{3}Information Technology University of the Punjab, Lahore, Pakistan\\[0.5em]
    \small\texttt{hasan.farooq@lhr.nu.edu.pk}, \texttt{murtaza.taj@lums.edu.pk}\\
    \small\texttt{mehwish.nasim@uwa.edu.au}, \texttt{arif.mahmood@itu.edu.pk}
}
\date{}

\begin{document}
\maketitle

\begin{abstract}
Medical Vision-Language Models (Med-VLMs) have demonstrated strong capabilities in clinical tasks. However, they often struggle to understand anatomical structures and spatial positioning, which are crucial for medical reasoning. To address this, we propose a \textbf{localization-aware} enhancement to the Med-VLM pipeline, introducing improvements at three levels: data, architecture, and alignment. First, we introduce \textbf{localization lens}, a set of expert-validated representations that provide richer anatomical and positional context. However, as these representations increase input complexity, we integrate pixel shuffle within the model architecture to filter and refine representations, enhancing spatial information processing while preserving anatomical continuity. Lastly, to effectively align the localization lens representations with textual features, we incorporate decoupled contrastive loss (DCL) alongside the standard loss function. This ensures better feature discrimination and robustness, particularly in data-limited medical settings. Through extensive evaluations on medical visual question answering (Med-VQA) datasets, we show that our methodology improves localization-driven performance across different Med-VLM architectures. Our analysis of localization-based questions further reveals that improvements in anatomy and spatial reasoning directly enhance the overall accuracy of Med-VQA upto 6.2\%. The proposed approach is \textbf{model-agnostic} and can be seamlessly integrated into existing Med-VLM pipelines. The dataset, code, and trained models will be made publicly available at \url{https://github.com/CVLABLUMS/localizationlens}.

\end{abstract}

\section{Introduction}

Medical Vision-Language Models (Med-VLMs) have improved medical image interpretation by integrating vision and text understanding into a unified framework \cite{med-vlm1, med-vlm2}. These models support various clinical applications, including automated report generation, medical visual question answering (Med-VQA), and clinical decision support \cite{med-vlm1, med-vlm2}. Many approaches adopt contrastive learning frameworks, such as CLIP, where image and text encoders are trained jointly on large-scale datasets \cite{med-adaption, pubmed, biomed}. More recently, large language models (LLMs) have been integrated into Med-VLMs to enable broader multimodal reasoning \cite{med-adaption, optional-3}. LLaVA-Med \cite{llava-med} fine-tunes a general-domain vision-language model on a figure-caption dataset with GPT-4-generated instruction data, while Med-Flamingo \cite{med-flamingo} is pre-trained on interleaved medical image-text pairs to improve few-shot generative VQA.

These contrastive and LLM-driven approaches demonstrate an increasing role of Med-VLMs in clinical reasoning and decision-making \cite{vlm-llm-1}. While these advancements have led to improvements in medical AI, Med-VLMs often rely on large architectures, requiring extensive computation and large datasets, thus limiting their accessibility \cite{med-adaption}. An alternative involves developing smaller Med-VLMs that retain key functionalities while reducing training and inference costs \cite{model-scaling, small-model}. However, since anatomical structures and spatial relationships are essential for accurate clinical reasoning \cite{Acosta2022, small-vlm1, med-moe}, smaller models often have difficulty encoding fine-grained localization cues, which are necessary for tasks such as disease detection, lesion identification, and reporting \cite{data-centric, token-crisis, model-scaling, med-moe}.

Several strategies have been explored to address these limitations, including data-centric methods, parameter-efficient scaling, advanced tokenization, region-based interpretations, and mixture-of-experts (MoE) approaches \cite{data-centric, token-crisis, model-scaling, med-moe}. For example, small language models (SLMs) have been proposed to improve reliability and accessibility in chronic disease management~\cite{small-model}, while Med-MoE \cite{med-moe} uses multiple small Med-VLMs as part of an MoE framework to improve Med-VQA performance. In the region-based interpretations, methodologies have been proposed to improve the localization capability of existing models by integrating bounding boxes within VLMs \cite{rgrg, chex}.

In this work, we explore the localization aspect in Med-VLMs by introducing a localization-aware enhancement that improves anatomical and spatial reasoning. Our approach consists of clinically meaningful representations, architectural modifications, and alignment refinements, making it model-agnostic and easily integrable into existing Med-VLMs. Our \textbf{contributions} are as follows:
\begin{itemize}
\item[(a)] We propose clinically meaningful representations, validated by experts at a Hospital \& Research Center, which act as \textbf{localization lens} to enhance anatomical and positional understanding in medical images.

\item[(b)] We integrate a \textbf{pixel-shuffle mechanism} within the model architecture to effectively handle the increased input complexity from the localization lens, improving the capture and refinement of spatial and anatomical details.

\item[(c)] We propose a \textbf{vision-language alignment} pipeline that first aligns the localization lens representations using decoupled contrastive learning, followed by their integration into Med-VQA tasks.

\item[(d)] We conduct a \textbf{localization analysis} to evaluate how the proposed localization-aware enhancements impact Med-VQA performance, demonstrating significant improvements in spatial reasoning and anatomical understanding across different Med-VLM architectures.

\end{itemize}

\section{Methodology}

\begin{figure}[h]
    \centering
    \includegraphics[width=0.99\textwidth]{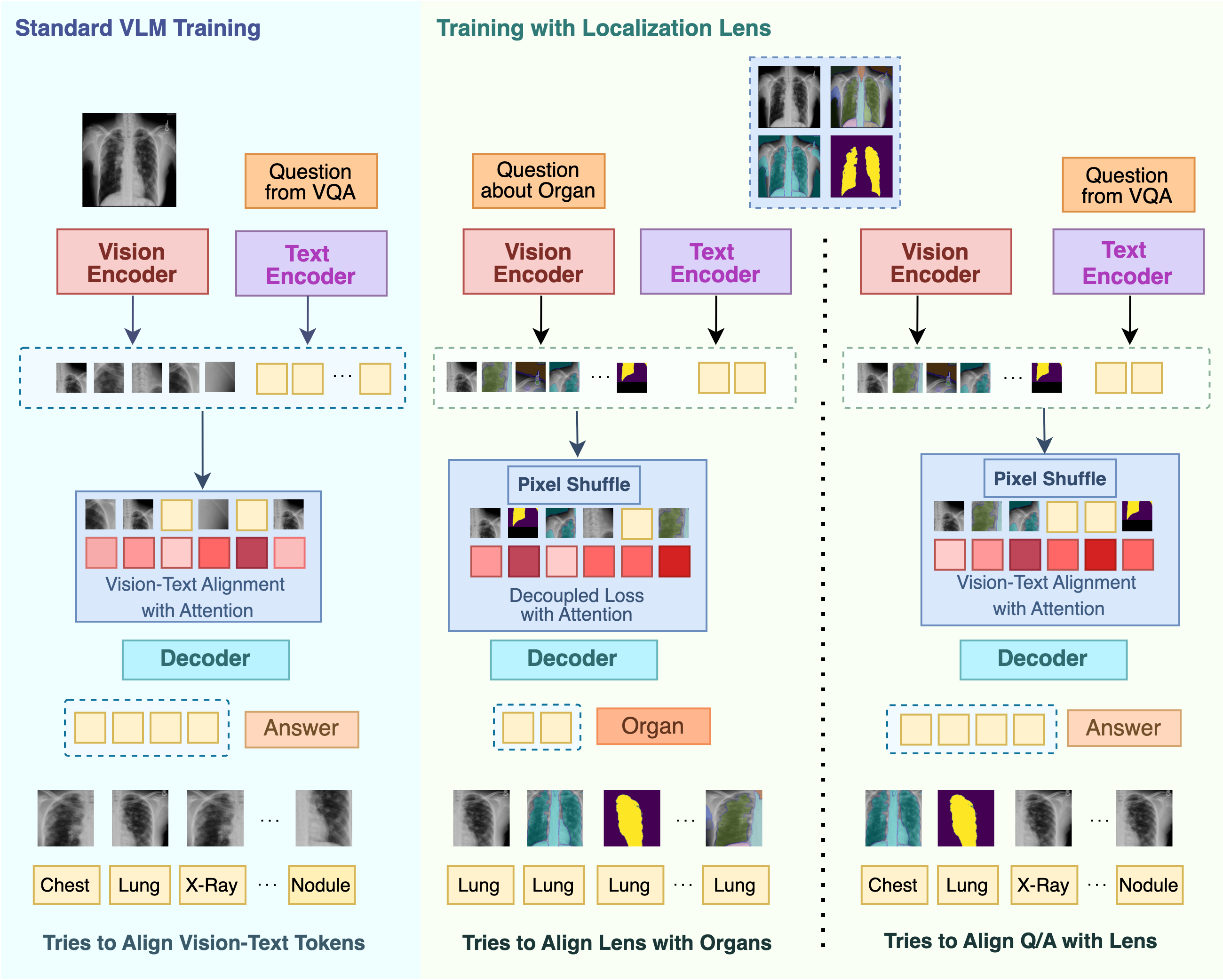}
    \caption{Comparison of the existing standard Med-VLM training with the proposed training utilizing localization lens.}
    \label{fig:architecture}
\end{figure}

\subsection{Preparation of Localization Lens}

The standard medical visual question answering (Med-VQA) datasets (such as VQA-RAD~\cite{vqa-rad} and SLAKE~\cite{slake}) come with a medical image, questions related to the image, answers to the questions, and metadata. In this work, on top of the public datasets, we constructed a clinically meaningful dataset by augmenting each clinical image with several complementary representations. These representations include (a) original image, (b) single-color segmented representation of original image, (c) multi-color coded segmented representation of original image, and (d) masked representation of the original image. We refer to these as \textbf{localization lens} as they enhance the localization context.

The quality and clinical relevance of these representations have been validated by experts at a Hospital and Research Center. The complete dataset will be made publicly available at \url{https://github.com/CVLABLUMS/localizationlens}.

\subsubsection{Uni and Multi-color Segmented Representations}
The segmented representations are obtained from the \textit{Segment Anything Model} (SAM) \cite{sam} and \textit{Med-SAM} \cite{med-sam1}. The representations from SAM produce coarser segmentation maps that may not be medically accurate and precise \cite{med-sam1, med-sam2}. However, these representations help in the generalization of the model as observed in our ablation studies (see Tab.~\ref{tab:ablation}). The representations from Med-SAM produce precise segmentation maps that capture anatomical structures and help the model learn the granular details of the medical image \cite{med-sam1}. Let $I$ be the original image, $f(\cdot)$ be the segmentation function, and $M_i$ be a segmentation mask where $1 \le i \le n$.

\begin{equation}
\{M_1, M_2, \ldots, M_n\} = f(I).
\end{equation}
Union of all segmentation masks results in full map:
\begin{equation}
\mathbf{M} = \{M_1 \cup M_2 \cup \ldots \cup M_n\}.
\end{equation}

After obtaining these segmentation masks, we blend these with the original image using single or multi color-coding. The blending is formulated as:

\begin{equation}
    I_{\text{blend}} = \alpha I + (1 - \alpha) \mathbf{C},
    \quad \text{where} \quad
    \mathbf{C} =
    \begin{cases}
        \mathbf{c} \sum_{i=1}^n M_i & _\text{(unicolor)} \\
        \sum_{i=1}^n M_i \mathbf{c}_i & _\text{(multicolor),}
    \end{cases}
\end{equation}

where $\alpha$ samples values from a normal distribution $0.50\pm 0.10$ for balanced blending and $\mathbf{c}_i\in \mathbb{R}^3 $ represents color for $i^th$ mask $M_i$.

\subsubsection{Masked Representations}

Each segment in the map represents an instance in the medical image. For example, in a chest X-Ray image, the question-answer pairs can be related to lungs. Masking the background regions assists the model to focus on learning the relevant features. We obtain these masked representations from segmentation maps with the help of experts. However, these masks can also be automatically obtained from medical segmentation models or medical vision language models. As a proof-of-concept, in our code, we share a VLM based promptable pipeline for automatically extracting these masks on other datasets (see Fig.~\ref{fig:architecture}).

 \subsection{Architectural Changes}

As shown in Fig.~\ref{fig:architecture}, the standard VLM training is optimized by aligning the image and text tokens. However, as we introduce context complexity by incorporating multiple representations, we optimize the architecture in two ways: (a) \textbf{input representation} and (b) \textbf{filtering representations}.

Instead of relying solely on an original image-text pair, our approach constructs an input that aggregates multiple visual representations along with the text.
For each original image-text pair, we randomly select two representations from (a) single-color segmentation maps, (b) multi-color segmentation maps, and (c) masked representations. These modalities are independently encoded by the vision encoder, and their embeddings are fused to form a unified representation.

\begin{equation}
E_{\text{fused}} = v(E_{\text{orig}}, E_{\text{rep1}}, E_{\text{rep2}})
\end{equation}
where \( E_{\text{orig}} \) is the embedding of the original image, and \( E_{\text{rep1}}, E_{\text{rep2}} \) are the embeddings of the selected representations. The function \( v \) represents the attention-based vision encoder mechanism.

Once the input is fused and partitioned into patches for processing, a \textbf{pixel shuffle mechanism} \cite{pixel-shuffle} is applied to refine the visual tokens. Unlike standard patch partitioning, which divides an image into fixed-size non-overlapping patches,

pixel shuffle takes an alternative approach by restructuring the spatial information into a more compact form. Given an input of shape \( (H, W, D \times r^2) \), pixel shuffle rearranges it into:

\begin{equation}
    T_{\text{out}}(h', w', d) = T\left(\frac{h'}{r}, \frac{w'}{r}, d \times r^2 + (h' \mod r) \times r + (w' \mod r) \right)
\end{equation}

which effectively reduces the number of patches while preserving local dependencies. This method provides two key advantages over conventional partitioning:

\begin{itemize}
    \item Instead of treating adjacent pixels independently, pixel shuffle redistributes sub-pixel information, reducing redundant tokenization. This helps the model filter relevant tokens for attention.
    \item Unlike uniform patch extraction, which can break meaningful structures, pixel shuffle retains continuity by preserving fine-grained details within each patch. This helps the training as the model focuses on patches from different representations.
\end{itemize}

By filtering representations more effectively, pixel shuffle enables the model to focus on semantically rich tokens, improving both computational efficiency and vision-text alignment.

\subsection{Training Details}
\label{sec:trdetails}
Since we introduce complexity by adding \textbf{localization lens}, our training is divided into two sequential phases designed to improve the vision-text alignment.

\textbf{Alignment of Visual Representations} The first phase serves as pre-training where we train the model for visual representations only. The organ represented in the image is used as text prompt, ensuring the encoder adapts to organ-specific features without interference from additional modalities. We employ \textbf{decoupled contrastive loss (DCL)} \cite{dcl}, which differs from standard InfoNCE by eliminating explicit negative repulsion. The InfoNCE loss is:

\begin{equation}
    L_{\text{InfoNCE}} = -\log \frac{\exp(\text{sim}(z_i, z_j) / \tau)}{\sum_{k} \exp(\text{sim}(z_i, z_k) / \tau)}
\end{equation}

where \( z_i, z_j \) are a positive pair, and negatives \( z_k \) force repulsion. In medical imaging, this can suppress meaningful variations, specifically when the data is scarce. DCL removes this constraint:

\begin{equation}
L_{DC} = -\log \frac{\exp(\langle \mathbf{z}_i^{(1)}, \mathbf{z}_i^{(2)} \rangle / \tau)}
{\exp(\langle \mathbf{z}_i^{(1)}, \mathbf{z}_i^{(2)} \rangle / \tau) + \sum_k \exp(\langle \mathbf{z}_i^{(1)}, \mathbf{z}_k \rangle / \tau)}
\end{equation}

where \( \mathbf{z}_i^{(1)} \) and \( \mathbf{z}_i^{(2)} \) are two augmented views of the same sample \( i \).

The model aims to maximize their similarity while handling negatives separately, avoiding repulsion. This improves:
(i) \textit{representation alignment}, as similar medical transformations remain close;
(ii) \textit{gradient stability}, avoiding conflicting updates; and
(iii) \textit{low-batch efficiency}, making it suitable for small datasets. This pre-training step enhances domain adaptation and prepares the encoder for multimodal fusion.

\textbf{Alignment of Question-Answer Pairs} In the second phase, we train the model for Med-VQA alignment. We use the actual question-answer pairs as text prompt along with the  \textbf{localization lens}. Since, we are training on the question-answer pairs, the standard \textbf{InfoNCE} contrastive loss (equation 6) is used for vision-text alignment. This penalizes the negative text tokens as well and ensures the alignment of visual representations with the question-answer (QA) pairs.

\subsection{Inference}

The localization lens are medically sound representations that enhance the anatomical and positional details whereas the proposed architectural design and alignment help the model extract patterns from the localization lens. After the model has been trained with the localization lens, we can perform vision-language tasks on the \textbf{original image} only.

\begin{table}[t]
    \centering
    \caption{Comparison of different Med-VQA models on the VQA-RAD and SLAKE.}
    \begin{tabular}{l c c c c | c}
        \toprule
        \textbf{Model} & \multicolumn{2}{c}{\textbf{VQA-RAD} \cite{vqa-rad}} & \multicolumn{2}{c}{\textbf{SLAKE} \cite{slake}} & \textbf{Model Size} \\
        & Open & Closed & Open & Closed &  \\
        \rowcolor{gray!20}
        \multicolumn{6}{c}{\textbf{Medical Vision Language Models}} \\
        RaDialog \cite{radialog} & 54.6 & 57.9 & 51.4 & 56.3 & 7B \\
        RaDialog (with Lens) & 55.3{\color{blue}\scriptsize$\uparrow$0.7} & 60.0{\color{blue}\scriptsize$\uparrow$2.1} & 53.4{\color{blue}\scriptsize$\uparrow$2.0} & 61.1{\color{blue}\scriptsize$\uparrow$4.8} & 7B \\
        \midrule
        Med-Flamingo \cite{med-flamingo} & 64.1 & 70.7 & 61.2 & 68.2 & 7B \\
        Med-Flamingo (with Lens) & 65.8{\color{blue}\scriptsize$\uparrow$1.7} & 72.1{\color{blue}\scriptsize$\uparrow$1.4} & 65.0{\color{blue}\scriptsize$\uparrow$3.8} & 71.6{\color{blue}\scriptsize$\uparrow$3.4} & 7B \\
        \midrule
        LLaVA-Med \cite{llava-med} & 61.2 & 76.2 & 70.4 & 75.0 & 7B \\
        LLaVA-Med (with Lens) & 62.4{\color{blue}\scriptsize$\uparrow$1.2} & 79.8{\color{blue}\scriptsize$\uparrow$3.6} & 74.0{\color{blue}\scriptsize$\uparrow$3.6} & 79.2{\color{blue}\scriptsize$\uparrow$4.2} & 7B \\
        \rowcolor{gray!20}
        \multicolumn{6}{c}{\textbf{Small Medical Vision Language Models}} \\
        Phi-MoE \cite{med-moe} & 36.7 & 61.8 & 43.9 & 57.0 & 3.6B \\
        Phi* (with Lens) & 48.3 & 56.7 & 51.3 & 58.4 & 2.7B \\
        \midrule
        SmolVLM \cite{smolvlm} & 53.8 & 57.6 & 46.5 & 55.9 & 1.7B \\
        SmolVLM (with Lens) & 58.0{\color{blue}\scriptsize$\uparrow$4.2} & 63.5{\color{blue}\scriptsize$\uparrow$5.9} & 54.2{\color{blue}\scriptsize$\uparrow$7.7} & 59.0{\color{blue}\scriptsize$\uparrow$3.1} & 1.7B \\
        \midrule
        SmolVLM \cite{smolvlm} & 48.3 & 51.4 & 43.2 & 49.8 & 0.5B \\
        SmolVLM (with Lens) & 53.6{\color{blue}\scriptsize$\uparrow$5.3} & 59.2{\color{blue}\scriptsize$\uparrow$7.8} & 49.3{\color{blue}\scriptsize$\uparrow$6.1} & 54.7{\color{blue}\scriptsize$\uparrow$4.9} & 0.5B \\
        \bottomrule
    \end{tabular}
    \label{tab:zero_shot_model_details}
    \par{\textbf{*} The results could not be reproduced, and we resorted to the base model.}
\end{table}

\section{Experiments}

\subsection{Medical VLMs}

We select 4 existing SOTA medical VLMs (RaDialog \cite{radialog}, Med-Flamingo \cite{med-flamingo}, LLaVA-Med \cite{llava-med}, and Med-MoE (Phi) \cite{med-moe}) and 2 non-medical VLMs (Phi \cite{phiii} and SmolVLM \cite{smolvlm} ]) for our experiments. We follow the training pipeline in Sec.~\ref{sec:trdetails} and use the VQA-RAD \cite{vqa-rad} and SLAKE \cite{slake} \textbf{datasets} for Med-VQA. To ensure a fair comparison, we follow a 70:30 train-test split.

In our experiments, we use 16-bit precision, learning rate of 0.001, weight decay of 0.01, and optimize using AdamW. For pixel-shuffle, we use an upscaling factor of 4. We report accuracy for closed-category questions and recall for open-category questions. Our results show that the performance of the Med-VLMs improves significantly with the proposed methodology (see Table~\ref{tab:zero_shot_model_details}).

From our results, we infer that the proposed methodology improves the performance of both the medical and non-medical VLMs. Interestingly, training models from scratch (i.e. non-medical models) shows more improvement. The possible explanation for this is the pre-training of VLMs that align the localization lens with organ labels more effectively \cite{pretraining}. Moreover, the results show that the small VLMs are capable of performance improvement with effective training pipelines and architectural design.

\subsection{Localization Analysis}

The Med-VQA datasets come with different types of Q/A pairs (i.e. disease identification, anatomical representation, positioning, modality, severity) \cite{vqa-rad, slake}. We conduct an extensive analysis on the Q/A pairs involving localization. We define localization as questions that (a) focus on presence of an anatomical structure or (b) focus on the positioning. We filter the questions that focus on the localization. Since there are thousands of Q/A pairs, we use the Llama-Vision (11B) \cite{llama} for filtering localization pairs.

\begin{table}[h]
    \centering
    \setlength{\tabcolsep}{5pt}

    \caption{Localization improvement results on combined dataset.}
    \begin{tabular}{lcc}
        \toprule
        \textbf{Model} & \textbf{Anatomical} & \textbf{Positioning} \\
        \midrule
        LLAVA-Med \cite{llava-med} & 4.7\% \(\uparrow\) & 5.6\% \(\uparrow\) \\
        Med-Flamingo \cite{med-flamingo} & 3.3\% \(\uparrow\) & 5.2\% \(\uparrow\) \\
        Phi* \cite{med-moe, phiii} & 2.8\% \(\uparrow\) & 3.6\% \(\uparrow\) \\
        SmolVLM \cite{smolvlm} (0.5B) & 4.2\% \(\uparrow\) & 5.8\% \(\uparrow\) \\
        SmolVLM \cite{smolvlm} (1.7B) & 3.7\% \(\uparrow\) & 6.1\% \(\uparrow\) \\
        \bottomrule
    \end{tabular}
    \label{tab:localization}
    \par{\textbf{*} The results could not be reproduced, and we resorted to the base model.}
\end{table}

Table~\ref{tab:localization} shows the results of our analysis on the combined VQA-RAD \cite{vqa-rad} and SLAKE \cite{slake} datasets with recall as the evaluation metric. Our results show that the proposed methodology significantly (upto $8.0\%$) improves the performance of Med-VLMs on the localization Q/A pairs. These results suggest that improvements in localization-based Q/A pairs contribute to the overall improvement in performance across different Med-VLMs.

\subsection{Ablation Studies}
To determine the contribution of each visual representation, we performed a series of ablation experiments. We compared the performance of models trained with only original images, original images with SAM \cite{sam} segmentation maps, original images with Med-SAM \cite{med-sam1} segmentation maps, original images with both segmentation maps, and original images with both segmentation sources plus masked images.
Our ablation study indicates that while both SAM \cite{sam} and Med-SAM \cite{med-sam1} improve results independently, their combined representations improves results by up to $6.2\%$. Also, although the (binary) masked representations alone do not represent meaningful structures, using these masks with segmentation maps achieves the best results.

\begin{table}[t]
    \centering

    \setlength{\tabcolsep}{3pt}
    \caption{Ablation study on VQA-RAD \cite{vqa-rad} with SmolVLM (1.7B).}
    \begin{tabular}{ccc|cc|c|cc}
        \toprule
        \multicolumn{3}{c|}{\textbf{Lens}} & \multicolumn{2}{c|}{\textbf{Loss}} & \textbf{Shuffle} & \multicolumn{2}{c}{\textbf{Metrics}} \tabularnewline
        \cmidrule(r){1-3} \cmidrule(r){4-5} \cmidrule(r){6-6} \cmidrule(r){7-8}
        \textbf{SAM} & \textbf{Med-SAM} & \textbf{Mask} & \textbf{InfoNCE} & \textbf{DCL} & \textbf{Pixel} & \textbf{Overall} & \textbf{Localization} \tabularnewline
        \midrule
        \xmark & \xmark & \xmark & \cmark & \xmark & \xmark & 55.4\% & - \tabularnewline
        \cmark & \xmark & \xmark & \xmark & \cmark & \cmark & 57.2\% & 4.9\% \(\uparrow\) \tabularnewline
        \xmark & \cmark & \xmark & \xmark & \cmark & \cmark & 56.8\% & 3.6\% \(\uparrow\) \tabularnewline
        \cmark & \cmark & \xmark & \xmark & \cmark & \cmark & 59.4\% & 7.4\% \(\uparrow\) \tabularnewline
        \midrule
        \cmark & \cmark & \cmark & \cmark & \xmark & \xmark & 57.3\% & 5.2\% \(\uparrow\) \tabularnewline
        \cmark & \cmark & \cmark & \cmark & \xmark & \cmark & 58.8\% & 6.2\% \(\uparrow\) \tabularnewline
        \cmark & \cmark & \cmark & \xmark & \cmark & \xmark & 57.7\% & 5.6\% \(\uparrow\) \tabularnewline
        \cmark & \cmark & \cmark & \xmark & \cmark & \cmark & 61.6\% & 8.0\% \(\uparrow\) \tabularnewline
        \bottomrule
    \end{tabular}
    \label{tab:ablation}
\end{table}

\section{Conclusion}

In this work, we integrate localization context as \textbf{localization lens}, a set of clinically meaningful representations, on the top of publicly available Med-VQA datasets. This localization lens adds both useful context and complexity for training the Med-VLMs. To address this, we propose integrating (a) \textbf{pixel-shuffle mechanism} within the architecture for filtering relevant context and (b) vision-text alignment with \textbf{decoupled contrastive loss}. These changes in the architecture design and vision-text alignment enable the Med-VLMs capture the context using localization lens effectively. Our experiments on Med-VQA datasets illustrate how the proposed methodology improves the localization context and subsequently, improves performance across Med-VLMs. The proposed methodology is \textbf{model-agnostic} and can be integrated into existing Med-VLM pipelines to improve their performance.

\section*{Competing interests}
The authors declare no competing interests relevant to the content of this article.

\bibliographystyle{plain}
\bibliography{references}

\end{document}